%% file: main.tex
\documentclass[runningheads,orivec]{llncs}

\usepackage[T1]{fontenc}

\usepackage{graphicx}
\usepackage{amsmath}
\usepackage{amssymb}
\usepackage{mathtools}
\usepackage{bm}
\usepackage{booktabs}
\usepackage{multirow}
\usepackage{xcolor}
\usepackage[compatibility=false]{caption}
\usepackage{subcaption}

\begin{document}

\title{Learning Representations from 3D Gaussian Splats}

\author{Julia Farganus \thanks{Equal contribution, authors listed alphabetically.} \and
Krzysztof \.{Z}urawicki \textsuperscript{\thefootnote}   \and
Arkadiusz Gawe\l{} \textsuperscript{\thefootnote} \and
\newline Weronika Jakubowska \and
Halina Kwa\'{s}nicka }

\authorrunning{J. Farganus, A. Gawe\l{}, K. \.{Z}urawicki et al.}

\institute{Department of Artificial Intelligence, Wroc\l{}aw University of Science and Technology, Wroc\l{}aw, Poland \\
\email{julia.farganus@student.pwr.edu.pl, krzysztof.zurawicki@pwr.edu.pl, weronika.jakubowska@pwr.edu.pl}}

\maketitle

\input{Section/abstract}

\input{Section/introduction}

\input{Section/related_work}

\input{Section/models}

\input{Section/experiments}

\input{Section/conclusions}

\input{Section/acknowledgements}

\bibliography{bibliography}
\bibliographystyle{splncs04}

\end{document}

%% file: Section/abstract.tex
\begin{abstract}

3D Gaussian Splatting (3DGS) is a recent approach for scene rendering. 
Although primarily designed for view synthesis, its potential for scene understanding tasks remains underexplored.
In this work, we conduct a comparative evaluation of various geometric deep learning architectures for the classification of 3D scenes represented using Gaussian Splatting. We benchmark point-based and graph-based models across both traditional point cloud datasets and dedicated Gaussian Splatting datasets. Scenes are embedded into latent representations, which are evaluated through end-to-end classification, linear probing, and clustering analysis.
Our study provides insight into the suitability of different geometry-aware architectures and input feature configurations for learning effective 3D Gaussian Splat representations. The results highlight consistent differences between architectural families and reveal the impact of Gaussian-specific attributes on the quality of representation.

\keywords{3D Gaussian Splatting  \and Latent representations \and Geometric Deep Learning.}

\end{abstract}

%% file: Section/introduction.tex
\section{Introduction}

As AI becomes increasingly integral to daily life, machines must not only process, but truly \textit{understand} data that describe the physical world. While natural phenomena often yield complex, high-dimensional data, research demonstrates \cite{Altman2018TheCO} that these representations can be compressed into fewer dimensions for downstream tasks. 
A primary example is 3D structures, whose unordered and sparse nature poses significant challenges for standard neural networks. 
To address these complexities, we focus on 3D point-based representations, exploring various architectural approaches to the classification task and analyzing their learned embedding space.
The importance of learning embeddings is grounded in the hypothesis that high-dimensional data actually resides on a lower-dimensional manifold \cite{Altman2018TheCO}. 
Specifically, within the 3D domain, expressive embeddings that preserve semantic distances and similarities can be utilized to bypass resource-intensive computations in the native 3D space.
Although creation of representations has been previously approached by masked autoencoders \cite{pang2022maskedautoencoderspointcloud} or multi-task learning \cite{hassani2019unsupervised}, we frame this task in a more direct manner, revisiting the efficacy of fully supervised learning.
Our primary focus lies in comparing the underlying data structures and connectivity patterns inherent to the selected architectures.

We base our research on the 3D Gaussian Splatting (3DGS) \cite{kerbl20233dgaussiansplattingrealtime} that has been up to date explored mostly for rendering and visualization. 
Crucially, like 3D point clouds, 3DGS consists of a set of unordered centroids, suggesting a natural compatibility with existing geometric neural architectures.
However, unlike raw points, these primitives exhibit spatial extent and directionality.
While this extended feature set is optimized strictly for visual rasterization rather than geometric fidelity -- and could theoretically introduce noise in downstream tasks -- pioneering work \cite{Zhang2025MACGS} has shown that incorporating auxiliary features like opacity, scale, and rotation can actually enhance shape recognition and yield more separable latent spaces.
Intuitively, incorporating purely visual features like opacity and color can significantly enhance class separation. 
For instance, textureless or solid objects are typically represented by a small number of large, opaque primitives with uniform color. 
Conversely, intricate objects with wire-like or porous structures require elongated primitives with varied opacities -- to account for holes -- and differing colors. 

Unlike purely centroid-based approaches, learning on 3DGS representations enables networks to differentiate between classes that share near-identical spatial geometry but possess distinct visual cues.
Motivated by the sparsity of research in this direction, we pose a fundamental question: \textit{Can neural networks successfully extract meaningful geometric representations from features originally optimized purely for visual fidelity}?
To systematically evaluate this, we model 3DGS scenes as graphs, allowing us to benchmark three distinct architectural paradigms: global connectivity, self-connectivity, and local connectivity (explained in Section \ref{subsec:models_chosen}).
We evaluate the quality of the resulting internal representations through a suite of downstream tasks, including linear probing alongside hard and soft clustering. 

This geometric interpretation aligns our framework directly with Geometric Deep Learning \cite{bronstein2021geometricdeeplearninggrids}.
This approach incorporates spatial symmetry and scale separation principles directly into neural architectures, effectively mitigating the curse of dimensionality. By applying graph embedding techniques, it maps sparse graphs with high-dimensional features into dense, low-dimensional vector spaces while strictly preserving structural properties.
In this work, we provide an evaluation of such embeddings, derived as a by-product of the classification task.
Our contributions are as follows:
\begin{enumerate}
    \item Comprehensive evaluation of embeddings through supervised classification and unsupervised clustering tasks.
    \item Establishment of Graph Neural Network baselines for classification on 3D Gaussian Splatting datasets.
    \item Quantitative benchmarking of state-of-the-art architectures with follow-up recommendations.
\end{enumerate}

%% file: Section/related_work.tex
\section{Related work}

Existing classification frameworks adopt one of three primary representation strategies: projection-based, voxel-based, and point-based methods, each offering a unique strategy to encode 3D geometric structures into latent embeddings.

\textbf{Point cloud} classification is a practical problem encountered in object detection, medicine, and 3D reconstruction \cite{ZHANG2023102456}.
The early solutions include projection-based methods \cite{su2015multiviewconvolutionalneuralnetworks,article_ptcld_3D_2d_comparison} and voxel-based methods \cite{zhou2017voxelnetendtoendlearningpoint,article_ptcld_3D_2d_comparison,Chen2023VoxelNeXtFS}.
Nevertheless, these methods ignore geometric representation and fail to learn recognizing abstract shapes.
More efficient solutions work directly with the unordered structure of 3D point cloud.
PointNet \cite{qi2017pointnetdeeplearningpoint} introduces an architecture that transforms each point independently.
Follow-ups include PointNet++ \cite{qi2017pointnetdeephierarchicalfeature}, which processes points hierarchically, and PointNeXt \cite{qian2022pointnext}, which extends PointNet++ by introducing residual connections and separable MLPs.

\textbf{Graph Neural Networks} extend point-based methods by explicitly modeling local and non-local relationships between points through graph structures.
Although many approaches are based on spectral graph theory, they struggle with generalization.
SplineCNN \cite{fey2018splinecnn} mitigates this limitation by operating directly in the spatial domain.
Its convolution operator aggregates node features using a trainable kernel function based on B-spline bases.
Another approach proposed in Graph Attention Network \cite{velickovic2018graph} (GAT) is to utilize masked attention layers, gaining efficiency from parallelization and shared weights for all edges.
Previously mentioned methods use static neighborhood that may hamper the ability to fuse information from distant nodes.
Dynamically built graphs were introduced in the Dynamic Graph Convolutional Neural Network \cite{wang2019dynamicgraphcnnlearning} (DGCNN), through an EdgeConv operation that uses feature space to find semantically similar elements, even though they are distant in the 3D coordinate space.
Other approaches for integrating contextual relationships involve pair-point impact function \cite{8954075} or superpoint graph \cite{Landrieu2017LargeScalePC}.
The work done in \cite{wu2020pointconvdeepconvolutionalnetworks}, \cite{xu2018spidercnndeeplearningpoint} addresses the extension of convolutional architectures to the 3D point cloud domain.

\textbf{Gaussian Splatting} \cite{kerbl20233dgaussiansplattingrealtime} is a novel technique for 3D visualization that has found use cases, for example, in medical imaging \cite{bonilla2024gaussianpancakesgeometricallyregularized3d} and autonomous driving \cite{11127564}. Let $\mathcal{G} = \{g_i\}_{i=1}^{N}$ denote a 3D Gaussian Splatting representation consisting of $N$ Gaussian primitives. According to the formulation \cite{kerbl20233dgaussiansplattingrealtime}, each primitive is characterized by its spatial position \(\mathbf{p}_i\), the anisotropic scaling factors  \(\mathbf{s}_i\), the orientation represented as a unit quaternion \(\mathbf{q}_i\), opacity \(\alpha_i\), and view-dependent color represented by spherical harmonic coefficients \(\mathbf{sh}_i\):
\[
  g_i = \left(\mathbf{p}_i, \mathbf{s}_i, \mathbf{q}_i, \alpha_i, \mathbf{sh}_i\right) \in \left(\mathbb{R}^{3}, \mathbb{R}^{3}, \mathbb{R}^{4}, \mathbb{R}^{1}, \mathbb{R}^{48}\right).
\]

In particular, considering only splat centroids \(\mathbf{p}_i\) makes them equivalent to a point cloud. As each feature carries a different meaning and lives in a separate domain, we also refer to them as \textit{heterogeneous attributes} later on.
The technique has been mainly employed for high quality rendering purposes with limited research exploring its potential as a replacement for point cloud representations.
Work done in \cite{zhang2025mitigating} shows that Gaussian Splatting objects perform better than point clouds on the 3D classification task.
Their enhanced representation mitigates ambiguities in distinguishing wire-like and flat surfaces, as well as transparent or reflective objects.
Gaussian-MAE \cite{ma2024shapesplat} is the first to perform self-supervised pretraining on a large-scale Gaussian Splatting dataset.
They show that pre-training on Gaussian Splats can encode shape priors that enhance classification performance on point cloud inputs.

\textbf{Current research} overlooks the potential of learned representations for 3D Gaussian Splats, largely due to the absence of dedicated mechanisms for sampling and aggregating splat features. To date, only \cite{ma2024shapesplat} has addressed this by introducing specialized grouping and pooling layers. However, their approach is intrinsically tied to a Transformer architecture and suffers from unbalanced feature weighting, limiting its generalization to standard point-based models. Furthermore, recent work by \cite{xin2025learningunifiedrepresentation3d} highlights the challenges of directly utilizing splat parameters due to their non-unique mapping and numerical heterogeneity.

In this work, we evaluate the impact of Gaussian features on the inherent mechanisms of Graph Neural Networks (GNNs) to assess their ability to mitigate these issues. We hypothesize that geometry-aware processing can yield robust embeddings suitable for downstream applications. Motivated by the multi-task learning approach in \cite{hassani2019unsupervised}, we evaluate the suitability of these embeddings for both supervised and unsupervised tasks. To our knowledge, no prior study has systematically examined the latent representations learned by these architectures in the domain of Gaussian Splatting. We also include point cloud architectures to compare different connectivity paradigms.

\begin{figure}[t]
\centering
\includegraphics[width=0.8\linewidth]{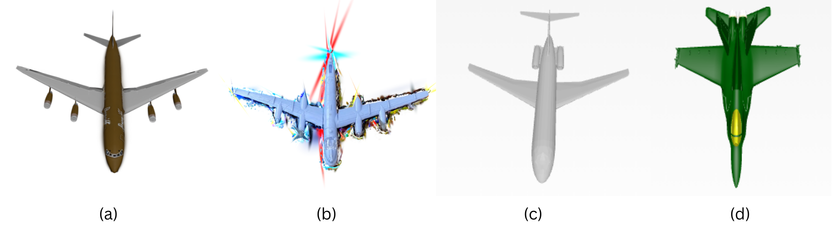}

\caption{Samples of \textit{airplane} class from datasets (a) ShapeSplat, (b) MACGS, (c) ModelNet40 and (d) ShapeNet. Gaussian Splatting representation provides is generally more detailed, but can suffer from artifacts as it's visible in (b). Mesh representations are smooth, but can't be directly processed by CNN's. }
\label{fig:samples_mosaique}
\end{figure}

%% file: Section/models.tex
\section{Used models and 3D Embeddings} 
We select several point-based methods and adapt their input layers and feature-processing mechanisms to handle the attributes of 3D Gaussian Splats. All models are trained end-to-end on classification tasks using both Gaussian Splatting and standard point cloud datasets, the latter serving as a baseline of uniformly sampled points -- a geometric distribution that cannot be reproduced by simply discarding the additional Gaussian Splat features. We then provide a brief explanation and categorization of the considered point-processing approaches.

\subsection{Approach Taxonomy} \label{subsec:models_chosen}

\paragraph{Global Connectivity.} Multi-Layer Perceptrons (MLPs) process flattened point clouds completely oblivious to their underlying geometric layout. In this setting, the network establishes global connectivity where each layer's output depends on the combination of all input points regardless of spatial proximity. Consequently, it lacks permutation invariance, meaning that reordering points or features drastically alters the output.

\paragraph{Self-Connectivity.} The PointNet family -- including PointNet, PointNet++, and PointNeXt -- processes individual points independently using shared weights, a~mechanism we define as self-connectivity. 
Permutation invariance is strictly enforced by aggregating these isolated point features via a symmetric pooling function (e.g., max-pooling). 
While the original PointNet operates globally, PointNet++ and PointNeXt introduce hierarchical grouping to capture expanding local neighborhoods, using relative distance encodings to improve spatial awareness.

\begin{figure}[t]
\centering
\includegraphics[width=0.8\linewidth]{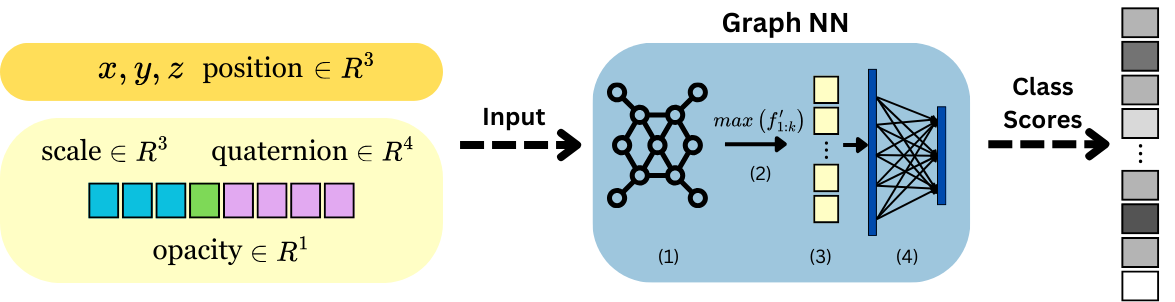}

\caption{\textbf{Gaussian Splatting processing via GNNs.} The input comprises spatial coordinates and a subset of auxiliary splat attributes (excluding spherical harmonics). The object is formulated as a graph and processed using the general \textit{message passing} paradigm (1). Global feature aggregation (2) is performed via max pooling, resulting in latent vector (3), followed by an MLP classifier (4).}
\label{fig:gnn_schema}
\end{figure}

\paragraph{Local Connectivity.} Graph Neural Networks (GNNs) -- specifically SplineCNN, DGCNN, and GAT -- explicitly model local neighborhoods by constructing edge relations among points. 
In this framework, each Gaussian primitive serves as a~node enriched with an extended feature vector, and local spatial relationships define the edges. This local connectivity allows for highly flexible graph construction; for instance, we define neighborhoods using $k=20$ nearest neighbors, which DGCNN dynamically updates in the latent feature space. 
Figure \ref{fig:gnn_schema} illustrates the general schema for these GNN architectures.

\subsection{Embeddings}

We formalize each network as a composition of a~feature extractor $h_{\theta}$ and a~classifier $g_{\gamma}$. While a~standard point cloud input is a set of 3D coordinates $V = \{v_i\}_{i=1}^n$ where $v_i = (x, y, z)$, the extended 3DGS input expands each primitive to a 14-dimensional vector $v_i = (x, y, z, s_{[1:3]}, q_{[1:4]}, \alpha, \text{sh}_{[1:3]})$ containing scaling, rotation, opacity, and color features. The final classification is computed as:
\begin{align*}
    \hat{y} = g_{\gamma} \left( \text{agg} \left( h_{\theta}(V) \right) \right),
\end{align*}
where $\text{agg}$ denotes a channel-wise max-pooling operator that aggregates the per-primitive feature matrix into a single vector. Crucially, we define the final \textit{scene embedding} as the fixed-size vector extracted immediately after this aggregation step (discarding $g_{\gamma}$). This choice captures the global structure synthesized by the encoder while allowing direct control over latent space expressiveness via the bottleneck dimension.

%% file: Section/experiments.tex
\section{Experiments}

Our experiments assess end-to-end classification performance across both Gaussian Splatting datasets and standard point cloud benchmarks. Furthermore, we evaluate the quality and discriminative power of the learned embeddings through downstream tasks.

\subsection{Datasets}
We selected two point clouds datasets ModelNet40 \cite{wu20153d} and ShapeNet \cite{chang2015shapenet}, as well as two Gaussian Splatting datasets: ShapeSplat \cite{ma2024shapesplat} and MACGS \cite{zhang2025mitigating}.
Although these datasets are semantically similar, containing common objects such as a~\textit{bag}, \textit{airplane}, or \textit{can} (see examples in Figure \ref{fig:samples_mosaique}), they differ significantly in structure.
This diversity allows us to evaluate the performance of the network in varying data qualities. Specifically, ShapeSplat was trained on 360-degree views, resulting in objects that are structurally complete and visually well-defined from all angles.
In contrast, MACGS relies on single-view supervision, leading to partial reconstructions where we anticipate decreased network performance. Finally, ModelNet40 and ShapeNet were converted to point cloud representations via uniform sampling from their original mesh formats.
 Regarding data set sizes, we operate on ModelNet40 -- 9,843 samples, 40 classes, ShapeSplat -- 30,843 samples, 55 classes, and MACGS -- 4,288 samples, 30 classes.
Additionally, we use ShapeNet-29 -- 16,318 samples, a~subset restricted to 29 classes for computational efficiency.

\begin{figure}[t]
\centering
\includegraphics[width=0.99\linewidth]{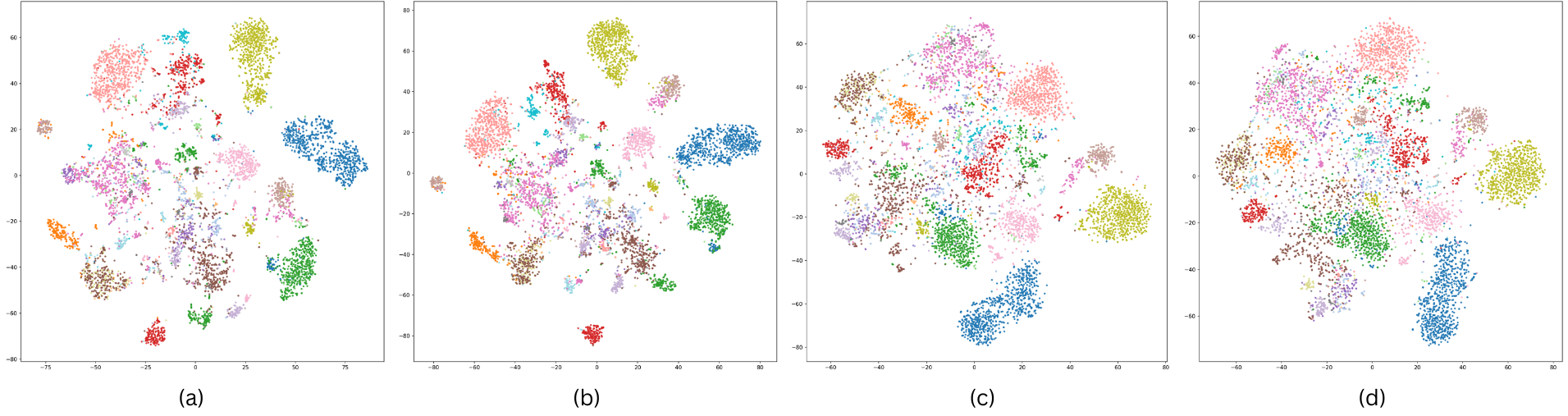}

\caption{\textbf{T-SNE visualization of ShapeSplat.} Images, from left to right present \textit{architecture}/\textit{features}: (a) PointNet++ / \textit{p}, (b) PointNet++ / \textit{psqo}, (c) Spline / \textit{p} and (d) Spline / \textit{psqo}. PointNet++ obtains visually more compact clusters (a), especially after including Gaussian splat features (b). On the other hand, Spline struggles to separate classes (c), with more observable spread when incorporating extra features (d).}
\label{fig:tsne_embeddings_shapesplat}
\end{figure}

\subsection{Experiments Setup}

\paragraph{Evaluation framework.} all models are trained end-to-end for the classification task without any explicit regularization on the embedding space. 
The quality of the learned representations is evaluated on a~series of downstream tasks, following the protocols in \cite{hassani2019unsupervised,ma2024shapesplat,chopin2024performance}. \\
\textbf{Linear probing} trains a~linear classifier on top of the frozen embeddings. The F1-score reveals whether the discriminative power originates from the classification head or the embeddings. \\
To evaluate the structure of the latent space, we utilize two clustering approaches.
\textbf{K-Means} clustering (hard clustering) uses \textit{Adjusted Mutual Information} (AMI) score to measure how well generated clusters align with the true class distribution, correcting for chance.
\textbf{Gaussian Mixture Models} (soft clustering) model each class as a~single multivariate Gaussian with a~diagonal covariance matrix. It provides insights into the structure of the latent space, under the intuition that semantic classes should map to compact clusters.
We evaluate the fit by the mean ($\mu_{\text{log}}$) and standard deviation ($\sigma_{\text{log}}$) of the log-likelihood of test set embeddings. 
To enable cross-dataset comparison independent of the absolute values of the mean log-likelihood ($\mu_{log}$) and its standard deviation ($\sigma_{log}$), we introduce the Mean Log Ratio, defined as:
\[
\text{Mean Log Ratio} = \frac{\sum_{i=1}^{C}\mu_{\text{log}}^{(i)}}{\sum_{i=1}^{C}\sigma_{\text{log}}^{(i)}}\,,
\]
where C denotes the number of classes. 

\begin{table}[t]
\centering
 \caption{\textbf{Model complexity and embedding dimensions.} \textit{P. (M)} denotes the number of parameters in millions; \textit{Emb. len.} refers to the embedding vector length. Parentheses indicate standard deviation.}
 \centering
    \begin{tabular}{|l|ccccccc|}
        \toprule
        Charact. & MLP & PointNet & PointNet++ & PointNeXt & DGCNN & Spline & GAT \\
        \midrule
        P. (M) & 2.50 {\scriptsize (0.30)} & 3.50 & 1.70 & 1.37 {\scriptsize (0.01)} & 1.85 {\scriptsize (0.25)} & 3.00 & 0.78 {\scriptsize (0.03)} \\
        \midrule
        Emb. len. & 1024 & 1024 & 1024 & 1024 & 1024 & 512 & 256 \\
        \bottomrule
    \end{tabular}
\label{tab:model_complexity_table}
\end{table}

\paragraph{Implementation Details.}
We train all models once, minimizing the standard Cross-Entropy loss using a~batch size of 32, except for GAT, where the batch size is reduced to 16 due to memory constraints imposed by the attention mechanism. Optimization is performed using AdamW with an initial learning rate of $10^{-3}$, a~weight decay of $10^{-4}$, and a~Cosine annealing scheduler. To avoid overfitting, we employ early stopping with a~patience of 3 epochs; the average training duration was approximately $12 \pm 2$ hours. The architecture implementations for PointNet and PointNet++ are adapted from \cite{PytorchPointnetPointnet2}, while PointNeXt follows the official codebase \cite{qian2022pointnext}. Graph-based models (SplineCNN, DGCNN, GAT) and the MLP baseline were constructed using geometric deep learning primitives from \cite{fey2019fast}. all experiments were conducted on a~single NVIDIA A100. Network sizes are given in Table \ref{tab:model_complexity_table}.

\paragraph{Data Preparation.} Input features underwent a~specific normalization, appropriate for each domain. Point coordinates were standardized to zero mean and unit variance. Gaussian opacities $o$ were mapped through a~sigmoid function, followed by standardization. To resolve antipodal ambiguity, quaternions were $L_{2}$-normalized and aligned to the upper hemisphere by multiplying with the sign of the real component. Scale features were standardized globally. ShapeSplat and MACGS point clouds were downsampled to $N=1024$ points via Farthest Point Sampling (FPS) \cite{qi2017pointnetdeephierarchicalfeature}. For mesh-based datasets (ShapeNet-29, ModelNet40), we first performed uniform surface sampling, followed by FPS to obtain a~canonical set of 1024 points. We use $80:20$ train/test split, with a~further $90:10$ train/validation split used during the training phase.

\paragraph{Points attributes.} For Gaussian Splats, we focus on comparing its two representations: centroids only (\textit{p}) and full splats which include position, scale (\textit{s}), quaternion (\textit{q}), and opacity (\textit{o}) attributes -- we denote the complete set as \textit{psqo}.
We also conduct an experiment with attributes exclusive to Gaussian Splats (\textit{sqo}) to assess their discriminative power in isolation.
With coordinate-only input, we utilize \textit{p} to represent position as feature attributes.
The color (expressed as spherical harmonics in 3DGS) is omitted, consistent with \cite{wang2019dynamicgraphcnnlearning} and \cite{zhang2025mitigating}, to ensure a~fair comparison with traditional point clouds.

\subsection{Results}

\subsubsection{Classification Performance.} accuracy results are detailed in Table \ref{tab:accuracy_results}.

\textit{MACGS dataset.} Spline achieves the best performance ($\approx 85\%$), competing closely with PointNet++. Incorporating full splat attributes (\textit{psqo}) generally improves performance over coordinates alone (\textit{p}). This gain is most prominent for PointNeXt ($+33.8$ pp), DGCNN ($+13.9$ pp) and GAT ($+13.3$ pp). Overall, PointNet, PointNet++, and Spline achieve the best performances (in the range of $\approx 76\text{-}84\%$), with PointNeXt, DGCNN, and GAT typically achieving $< 70\%$. MLP, as expected, yields the worst results. 

\textit{ShapeSplat dataset.} The best performance is achieved by similarly scoring Spline, PointNet, and PointNet++ in the range of $84\text{-}86\%$. DGCNN performs better than PointNeXt, scoring $\approx 79\text{-}83\%$, which is a~notable improvement over the MACGS results. a~similar enhancement is observed for GAT with respect to the corresponding experiments on MACGS. For PointNeXt and GAT, we observe a~noticeable improvement ($+10$ pp) over the coordinate-only setting (\textit{p}). MLP, even though scoring behind, improves with the enhanced quality of the splats.

\textit{Point Cloud Benchmarks.} On ModelNet40 and ShapeNet-29, PointNet++ achieves the highest accuracy, closely followed by Spline and PointNet. although ModelNet40 results on PointNet and PointNeXt (72-78\%) are slightly lower than reported in \cite{qian2022pointnext} this is likely due to the specific split or FPS preprocessing. Notably, PointNeXt exhibits moderate degradation compared to the leading architectures on ShapeNet-29, while for GAT and DGCNN, the performance gap is significant. While the MACGS dataset typically provides lower accuracy than point cloud benchmarks, PointNet, Spline, and GAT exhibit improved performance relative to ModelNet40. In contrast, the ShapeSplat dataset provides distinct gains, most prominently for DGCNN and GAT, while PointNet and Spline also surpass their ModelNet40 baselines.

\begin{table}[ht]
\centering
\caption{\textbf{Accuracy results for end-to-end training.} Numbers in bold denote the best result within a~given dataset and feature. Underlined numbers indicate the best result within a~given dataset and architecture. although accuracy does not reflect class imbalance, it remains a~common metric reported in related studies such as \cite{zhang2025mitigating}, \cite{wang2019dynamicgraphcnnlearning}, and \cite{ma2024shapesplat}}
\begin{tabular}{|ll|ccccccc|}
        \toprule
        Dataset & Attr. & MLP & PointNet & PointNet++ & PointNeXt & DGCNN & Spline & GAT
        \\
        \midrule
        
        ModelNet40 & \textit{p} & 4.052 & 78.201 & \textbf{86.791} & 72.639 & 75.810 & 80.713 & 40.438 \\
        \cline{1-9}
        ShapeNet-29 & \textit{p} & 19.761 & 86.737 & \textbf{89.130} & 66.796 & 3.639 & 88.276 & 11.602 \\
        \cline{1-9}
        
        \multirow{3}{*}{MACGS} & \textit{p} & 14.510 & 77.170 & \textbf{79.760} & 35.016 & 57.024 & 78.928 & 46.858 \\
         & \textit{psqo} & \underline{23.660} & \underline{79.570} & \underline{84.658} & 68.806 & \underline{70.980} & \underline{\textbf{84.935}} & \underline{60.166} \\
         & \textit{sqo} & 13.216 & 76.250 & \textbf{82.348} & \underline{69.850} & 51.017 & 78.281 & 42.144 \\
        \cline{1-9}
        \multirow{3}{*}{ShapeSplat} & \textit{p} & 40.901 & \underline{85.730} & 86.323 & 57.208 & 80.951 & \underline{\textbf{86.685}} & 68.268 \\
         & \textit{psqo} & \underline{56.658} & 85.729 & \underline{\textbf{86.504}} & \underline{67.678} & \underline{83.611} & 85.794 & \underline{80.021} \\
         & \textit{sqo} & 42.942 & \textbf{85.006} & 84.773 & 63.711 & 79.259 & 84.915 & 70.657 \\
        \cline{1-9}
        \bottomrule
    \end{tabular}
    
\label{tab:accuracy_results}
\end{table}

\begin{table}[ht]
\centering
\caption{\textbf{Linear probing F1 Score.} Numbers in bold denote the best result within a~given dataset and a~feature. Underlined numbers indicate the best result within a~given dataset and an architecture.}
\label{tab:f1-score_results}
\begin{tabular}{|ll|ccccccc|}
\toprule
 Dataset & Attr. & MLP & PointNet & PointNet++ & PointNeXt & DGCNN & Spline & GAT \\
\midrule
ModelNet40 & \textit{p} & 0.316 & 81.418 & \textbf{85.380} & 76.322 & 53.331 & 54.181 & 42.810 \\
\cline{1-9}
ShapeNet-29 & \textit{p} & 6.521 & 86.366 & \textbf{86.682} & 79.198 & 6.521 & 69.338 & 12.521 \\
\cline{1-9}
\multirow{3}{*}{MACGS} & \textit{p} & 3.355 & \textbf{78.091} & 76.247 & 31.724 & 19.555 & \underline{44.004} & 14.928 \\
 & \textit{psqo} & \underline{5.456} & \underline{79.674} & \underline{\textbf{81.317}} & 60.076 & \underline{20.333} & 39.303 & \underline{18.162} \\
 & \textit{sqo} & 4.729 & 77.091 & \textbf{80.932} & \underline{63.119} & 11.498 & 33.945 & 13.668 \\
\cline{1-9}
\multirow{3}{*}{ShapeSplat} & \textit{p} & 6.001 & \textbf{84.868} & 83.988 & 77.349 & \underline{43.153} & \underline{62.050} & 15.355 \\
 & \textit{psqo} & \underline{6.958} & \underline{\textbf{85.412}} & \underline{84.104} & \underline{79.925} & 42.512 & 61.034 & \underline{20.506} \\
 & \textit{sqo} & 6.630 & 81.961 & \textbf{82.180} & 77.370 & 34.482 & 54.008 & 19.069 \\
\cline{1-9}
\bottomrule
\end{tabular}
\end{table}

\subsubsection{Linear Probing.}
We report the weighted average F1-score to account for class imbalance. The results are summarized in Table \ref{tab:f1-score_results}.

\begin{figure}[t]
\begin{subfigure}{0.48\textwidth}
  \centering
  \includegraphics[width=1.0\linewidth]{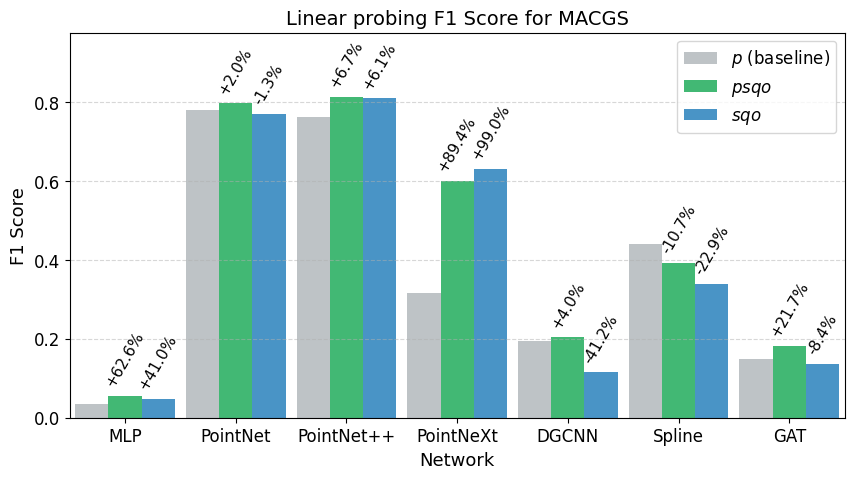}
  \caption{\textbf{Linear probing - MACGS}}
  \label{fig:macgs_linear_probing}
\end{subfigure}
\hfill
\begin{subfigure}{0.48\textwidth}
  \centering
  \includegraphics[width=1.0\linewidth]{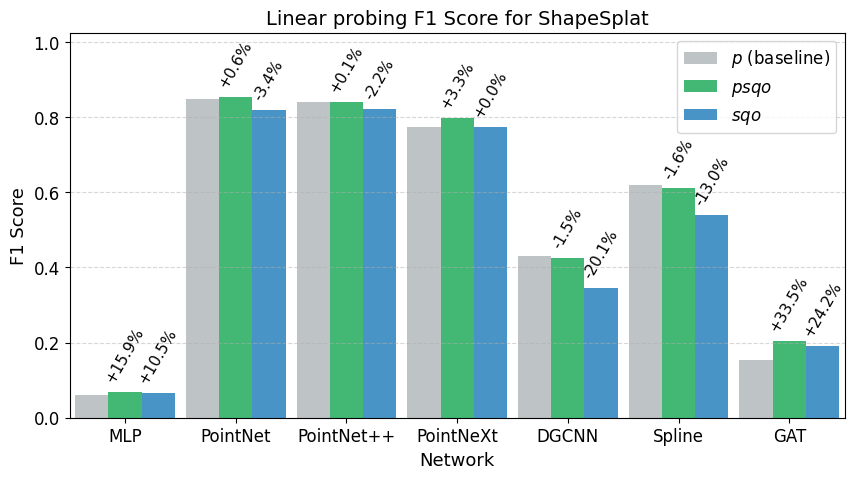}
  \caption{\textbf{Linear probing - ShapeSplat}}
  \label{fig:shapesplat_linear_probing}
\end{subfigure}
\caption{\textbf{Linear probing F1 Score} for each feature configuration. Percentage value above the bar containers denote the relative gain or lost when using \textit{p}.}
\label{fig:linear_probing}
\end{figure}

\textit{MACGS dataset.} While PointNet and PointNet++ achieve robust performance ($\approx80\%$), PointNeXt exceeds the GNN baselines only when using the \textit{psqo} and \textit{sqo} features ($\approx 60\%$). among GNNs, Spline yields the best results, yet it attains only half the accuracy of the PointNet family. DGCNN and GAT score is only 20\%, indicating a~severe loss of generalization compared to their end-to-end accuracy. Incorporating Gaussian attributes generally drives improvement; as shown in Figure \ref{fig:macgs_linear_probing}, PointNeXt exhibits a~relative improvement over 90\%, with \textit{sqo} slightly outperforming $psqo$. Conversely, Spline is the only architecture that suffers a~relative regression ($\approx 10\%$) with Gaussian features.

\textit{ShapeSplat dataset.} The PointNet family outperforms GNN-based architectures, showing greater robustness to class imbalance. 
PointNet and PointNet++ achieve 81–85\% accuracy, with PointNeXt close behind $\approx 77\text{-}79\%$.
Spline exhibits a~substantial drop relative to its fully supervised performance, lagging more than 20 percentage points behind the PointNet family. 
DGCNN and GAT perform poorly, although DGCNN roughly doubles its MACGS performance ($\approx 20\%$ and $\approx 40\%$ respectively). 
As shown in Figure~\ref{fig:shapesplat_linear_probing}, Gaussian features typically yield modest gains (1–2 pp), but for DGCNN and Spline, the coordinate-only input (\textit{p}) performs slightly better. 
Overall, the results indicate that GNNs may struggle to model interactions between heterogeneous feature types.

\textit{Point Cloud Benchmarks.} On point cloud benchmarks, PointNet++ consistently secures the highest scores, closely followed by PointNet and PointNeXt. 
In contrast, GNNs exhibit notable degradation, often yielding near-random results (with a~minor exception for Spline on ShapeNet-29).
Compared to the ModelNet40 baseline, the MACGS dataset exhibits a~performance drop across all architectures, with the degradation being most visible in GNNs. In the case of ShapeSplat, however, we observe improvements for PointNet, PointNeXt, and Spline. While the 3DGS format does not outperform the ShapeNet-29 baseline for the PointNet family, it notably enhances the performance of GAT and DGCNN architectures. 

\subsubsection{Clustering.} 
This evaluation assesses the geometric properties of the embedding space, specifically focusing on intra-class compactness and inter-class separability. The results are provided in Tables \ref{tab:ami_results} and \ref{tab:mean_log_ratio_results}.

\begin{table}[ht]
\centering
\caption{\textbf{Clustering - AMI results}. Numbers in bold denote the best result within a~given dataset and a~feature. Underlined numbers indicate the best result within a~given dataset and an architecture.}
\label{tab:ami_results}
\begin{tabular}{|ll|ccccccc|}
\toprule
 Dataset & Attr. & MLP & PointNet & PointNet++ & PointNeXt & DGCNN & Spline & GAT \\
\midrule
ModelNet40 & \textit{p} & 0.000 & 0.650 & \textbf{0.665} & 0.487 & 0.371 & 0.569 & 0.339 \\
\cline{1-9}
ShapeNet-29 & \textit{p} & 0.006 & 0.558 & \textbf{0.643} & 0.494 & 0.006 & 0.626 & 0.008 \\
\cline{1-9}
\multirow{3}{*}{MACGS} & \textit{p} & 0.008 & \textbf{0.619} & 0.510 & 0.223 & 0.148 & \underline{0.294} & 0.085 \\
 & \textit{psqo} & \underline{0.016} & \underline{\textbf{0.651}} & \underline{0.527} & \underline{0.430} & \underline{0.152} & 0.235 & \underline{0.096} \\
 & \textit{sqo} & 0.012 & \textbf{0.633} & 0.509 & 0.370 & 0.137 & 0.209 & 0.076 \\
\cline{1-9}
\multirow{3}{*}{ShapeSplat} & \textit{p} & 0.017 & 
\textbf{0.732} & 0.679 & 0.603 & 0.601 & \underline{0.645} & \underline{0.273} \\
 & \textit{psqo} & \underline{0.024} & \underline{\textbf{0.735}} & \underline{0.711} & \underline{0.640} & \underline{0.643} & 0.620 & 0.244 \\
 & \textit{sqo} & 0.016 & 0.648 & \textbf{0.675} & 0.594 & 0.479 & 0.603 & 0.167 \\
\cline{1-9}
\bottomrule
\end{tabular}
\end{table}

\begin{table}[ht]
\centering
\caption{\textbf{Clustering - Mean Log Ratio results}. Numbers in bold denote the best result within a~given dataset and a~feature. Underlined numbers indicate the best result within a~given dataset and an architecture. Results missing for ModelNet40 and ShapeNet-29 come from the fact that the GMM may fail to converge. }
\label{tab:mean_log_ratio_results}
\begin{tabular}{|ll|ccccccc|}
\toprule
 Dataset & Attr. & MLP & PointNet & PointNet++ & PointNeXt & DGCNN & Spline & GAT \\
\midrule
ModelNet40 & \textit{p} & - & 2.219 & \textbf{2.965} & 2.039 & 1.794 & 2.337 & 2.218 \\
\cline{1-9}
\multirow[t]{3}{*}{ShapeNet-29} & \textit{p} & - & 2.253 & 2.609 & \textbf{3.385} & - & 2.245 & 1.961 \\
\cline{1-9}
\multirow{3}{*}{MACGS} & \textit{p} & 1.454 & \textbf{2.755} & \underline{2.577} & 1.905 & \underline{2.079} & 1.729 & \underline{1.734} \\
 & \textit{psqo} & 1.672 & 2.390 & 2.055 & 2.109 & 2.008 & \underline{\textbf{2.397}} & 1.581 \\
 & \textit{sqo} & \underline{1.841} & \underline{\textbf{2.846}} & 2.170 & \underline{2.250} & 1.618 & 1.841 & 1.720 \\
\cline{1-9}
\multirow{3}{*}{ShapeSplat} & \textit{p} & 1.704 & \underline{\textbf{2.889}} & 2.803 & \underline{2.442} & 1.764 & 1.946 & 1.656 \\
 & \textit{psqo} & 1.725 & 2.575 & \textbf{2.708} & 2.419 & \underline{2.015} & \underline{2.112} & \underline{1.950} \\
 & \textit{sqo} & \underline{1.732} & 2.377 & \underline{\textbf{2.954}} & 1.593 & 1.578 & 1.820 & 1.889 \\
\cline{1-9}
\bottomrule
\end{tabular}
\end{table}

\begin{figure}[t]
\centering
\includegraphics[width=0.9\linewidth]{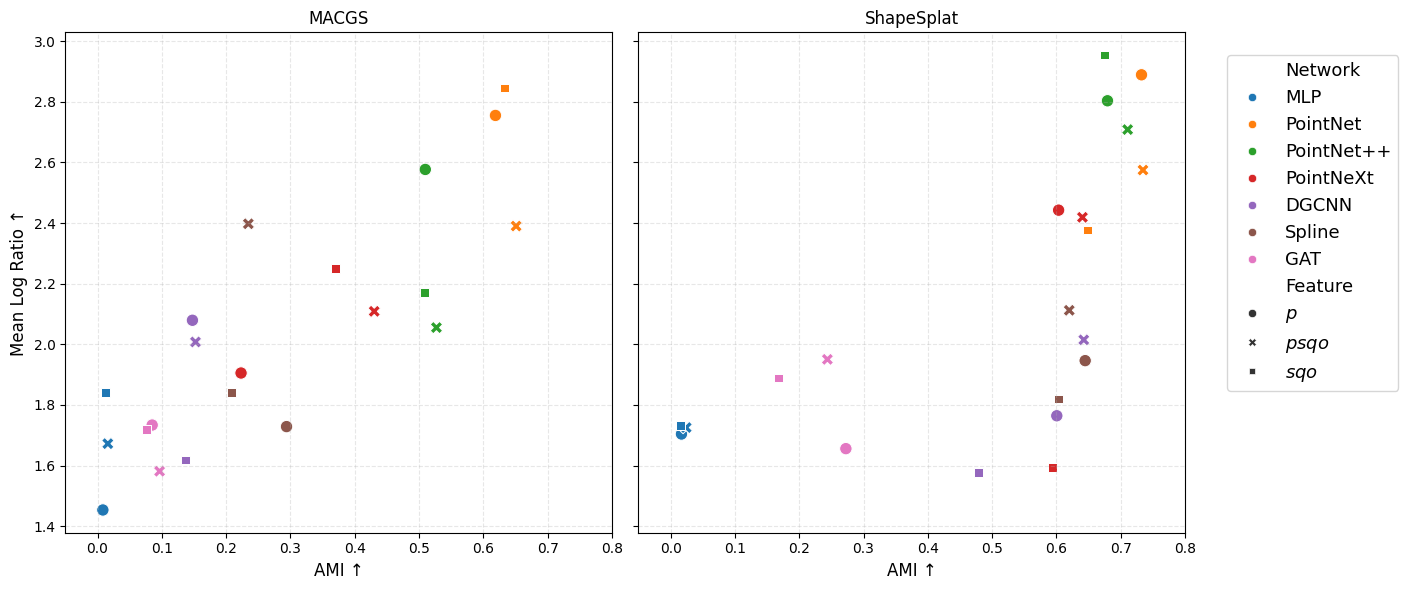}

\caption{\textbf{Clustering Evaluation}. Relation between AMI from k-Means and standard deviation of mean Log Probabilities from GMM. Marker colors represent a~unique architecture, and their style a~given feature.}
\label{fig:clustering_evaluation}
\end{figure}

\textit{MACGS dataset.} PointNet achieves the highest AMI ($\approx 0.6$), followed by PointNet++ ($\approx 0.5$), while PointNeXt, Spline, and DGCNN obtain suboptimal scores ($\approx 0.2\text{-}0.4$), with GAT and MLP results being negligible. Incorporating splat attributes (\textit{psqo}) generally enhances AMI -- most notably for PointNeXt (+0.2) -- yet this does not necessarily translate to improved Log Ratio. Specifically, Spline is the only architecture where adding Gaussian features to positions improves the Log Ratio; conversely, PointNet and PointNeXt peak using standalone Gaussian features (\textit{sqo}), while PointNet++, DGCNN, and GAT favor coordinate-only input (\textit{p}). as illustrated in Figure \ref{fig:clustering_evaluation}, there is a~trade-off between these metrics: PointNet (\textit{sqo} and \textit{psqo}) defines the Pareto frontier, whereas for PointNeXt, relying solely on coordinates degrades both metrics. Meanwhile, Spline with full feature set (\textit{psqo}) offers a~favorable compromise, sacrificing negligible AMI for a~significantly improved Log Ratio compared to the baseline.

\textit{ShapeSplat dataset.} PointNet and PointNet++ dominate the clustering benchmarks, achieving high AMI scores for coordinate-based (\textit{p}) and full (\textit{psqo}) configurations ($\approx 0.7$), with only minor degradation observed for the \textit{sqo} subset. DGCNN, Spline, and PointNeXt follow closely, exceeding an AMI threshold of 0.6 in the best case. although integrating splat attributes generally yields performance gains, except for Spline and GAT, relying solely on Gaussian features (\textit{sqo}) typically causes regression. In terms of Log Ratio, PointNet and PointNet++ score top values, similarly as with AMI results. Interestingly, for the PointNet architectures, augmenting positions with Gaussian features (\textit{psqo}) provides no significant benefit over the baseline (\textit{p}), although PointNet++ notably achieves its peak Log Ratio using \textit{sqo} alone. Conversely, GNNs demonstrate a~clear dependency on Gaussian attributes to surpass the MLP baseline.
As visualized in Figure \ref{fig:clustering_evaluation}, PointNet and PointNet++ occupy the optimal upper-right quadrant, striking a~balance between cluster quality and probabilistic fit.
In contrast, PointNeXt and DGCNN exhibit degradation in the \textit{sqo} setting, while GAT remains suboptimal, despite improving its MACGS performance.
Overall, ShapeSplat yields consistently robust AMI scores across most architectures, with the primary optimization challenge lying in the Log Ratio. We provide a~visual assessment of the learned embedding space in Figure \ref{fig:tsne_embeddings_shapesplat}. 

\textit{Point Cloud Benchmarks.} For both datasets, PointNet++, PointNet and Spline attain good performance for both clustering types. For DGCNN and GAT we can observe a~degradation, especially for ShapeNet-29, remaining near-zero scores obtained by MLP baseline. On the MACGS dataset, clustering performance generally degrades compared to point cloud baselines. In contrast, ShapeSplat achieves improvements for PointNet and DGCNN, as well as for PointNet++ and PointNeXt, although with some regressions in the Mean Log Ratio metric.

%% file: Section/conclusions.tex
\section{Conclusions}

Contrary to initial intuition, GNNs generally struggle with end-to-end classification, with the notable exception of SplineCNN, which retains accuracy comparable to the PointNet family. 
However, these networks suffer from significant performance drops in linear probing and clustering, suggesting that standard message-passing mechanisms require adaptation for the Gaussian Splat domain.
Since Gaussian Splats are non-uniformly distributed in space (in contrast to uniformly sampled point clouds), the constructed k-NN neighborhoods may fail to reflect semantically relevant connections.
Examining the architectures, DGCNN appears to struggle with consistently capturing local geometric structure due to the dynamic recomputation of the k-NN graph at every layer.
As it computes distances in feature space, the heterogeneity of attributes likely hinders the learning of meaningful connections.
Conversely, GAT operates on fixed neighborhoods but discards explicit structural information in favor of a~masked attention mechanism.
Additionally, it aggregates features via linear combinations that, as previously hypothesized, may be suboptimal for this data type. SplineCNN employs local processing mechanisms that closely resemble the operations of Convolutional Neural Networks (CNNs).
Its weights are locally constrained within a~defined kernel, operating directly on relative distances. This structural inductive bias likely explains its classification performance, although its generalization capability degrades significantly in downstream tasks compared to the PointNet family. As anticipated, the global processing schema of the MLP consistently provides the lowest scores; however, the use of 3DGS representations improves performance compared to point cloud baselines.

\textit{Future directions.} We argue that incorporating dedicated message-passing mechanisms could greatly benefit the Gaussian Splat representation, as each input feature carries a~distinct semantic meaning.
Although some progress has been made in this direction through Splat Grouping and Pooling layers \cite{ma2024shapesplat}, it still does not enhance the mathematical interpretation and possible interactions between features.
Furthermore, the current standard for point cloud preprocessing utilizes Farthest Point Sampling (FPS).
Since FPS considers only spatial distance, it may discard splats that contain auxiliary features critical to object representation.
Consequently, we suggest that considering the probabilistic interpretation of Gaussian Splats could drive the implementation of a~specialized, feature-aware sampling routine.
Finally, the robust, generalizable performance of the PointNet family in our experiments underscores the need to maintain a~balance between self- and local connectivity in future architectural designs.

%% file: Section/acknowledgements.tex
\begin{credits}

\subsubsection{\ackname}
The work of W. Jakubowska was supported by the National Centre of Science (Poland) Grant No. 2023/50/E/ST6/00068.
We gratefully acknowledge the Polish high-performance computing infrastructure PLGrid (HPC Center: ACK Cyfronet AGH) for providing computational facilities and support under computational grant No. PLG/2025/018862. Authors would like to acknowledge Maciej Zi\k{e}ba for review and his insightful comments on paper.

\subsubsection{\discintname}
The authors have no competing interests to declare that are relevant to the content of this article.

\end{credits}